\documentclass[a4paper,conference]{IEEEtran}
\makeatletter
\def\endthebibliography{%
	\def\@noitemerr{\@latex@warning{Empty `thebibliography' environment}}%
	\endlist
}
\makeatother
\ifCLASSINFOpdf
\else
\fi

\usepackage[utf8]{inputenc} %unicode support
\usepackage{graphicx}
\usepackage{times}
\usepackage{textcomp}  
\usepackage{cite}                      % needed to automatically sort the references
\usepackage{tabu}                      % only used for the table example
\usepackage{booktabs}                  % only used for the table example
\usepackage{mathtools}
\usepackage{amsmath,amsfonts,amssymb}
\usepackage{adjustbox}
\usepackage{latexsym}
\usepackage{url}
\usepackage{xcolor}
\usepackage{multirow}
\usepackage{subfig}
\DeclareMathOperator*{\E}{\mathbb{E}}

\begin{document}
%
% paper title
% Titles are generally capitalized except for words such as a, an, and, as,
% at, but, by, for, in, nor, of, on, or, the, to and up, which are usually
% not capitalized unless they are the first or last word of the title.
% Linebreaks \\ can be used within to get better formatting as desired.
% Do not put math or special symbols in the title.
\title{Neural Architecture Search for Image Super-Resolution Using Densely Constructed Search Space: DeCoNAS}

% author names and affiliations
% use a multiple column layout for up to three different
% affiliations
\author{\IEEEauthorblockN{Joon Young Ahn}
\IEEEauthorblockA{Dept. of Electrical and Computer Engineering, \\ INMC, Seoul National University\\
Seoul, Korea \\ Email: sohwa360@ispl.snu.ac.kr}
\and
\IEEEauthorblockN{Nam Ik Cho}
\IEEEauthorblockA{Dept. of Electrical and Computer Engineering, \\ INMC, Seoul National University\\
	Seoul, Korea \\ Email: nicho@snu.ac.kr}}

% conference papers do not typically use \thanks and this command
% is locked out in conference mode. If really needed, such as for
% the acknowledgment of grants, issue a \IEEEoverridecommandlockouts
% after \documentclass

% for over three affiliations, or if they all won't fit within the width
% of the page, use this alternative format:
%
%\author{\IEEEauthorblockN{Michael Shell\IEEEauthorrefmark{1},
%Homer Simpson\IEEEauthorrefmark{2},
%James Kirk\IEEEauthorrefmark{3},
%Montgomery Scott\IEEEauthorrefmark{3} and
%Eldon Tyrell\IEEEauthorrefmark{4}}
%\IEEEauthorblockA{\IEEEauthorrefmark{1}School of Electrical and Computer Engineering\\
%Georgia Institute of Technology,
%Atlanta, Georgia 30332--0250\\ Email: see http://www.michaelshell.org/contact.html}
%\IEEEauthorblockA{\IEEEauthorrefmark{2}Twentieth Century Fox, Springfield, USA\\
%Email: homer@thesimpsons.com}
%\IEEEauthorblockA{\IEEEauthorrefmark{3}Starfleet Academy, San Francisco, California 96678-2391\\
%Telephone: (800) 555--1212, Fax: (888) 555--1212}
%\IEEEauthorblockA{\IEEEauthorrefmark{4}Tyrell Inc., 123 Replicant Street, Los Angeles, California 90210--4321}}

% use for special paper notices
%\IEEEspecialpapernotice{(Invited Paper)}

% make the title area
\maketitle

% As a general rule, do not put math, special symbols or citations
% in the abstract
\begin{abstract}
The recent progress of deep convolutional neural networks has enabled great success in single image super-resolution (SISR) and many other vision tasks. Their performances are also being increased by deepening the networks and developing more sophisticated network structures. However, finding an optimal structure for the given problem is a difficult task, even for human experts. For this reason, neural architecture search (NAS) methods have been introduced, which automate the procedure of constructing the structures. In this paper, we expand the NAS to the super-resolution domain and find a lightweight densely connected network named DeCoNASNet. We use a hierarchical search strategy to find the best connection with local and global features. In this process, we define a complexity-based penalty for solving image super-resolution, which can be considered a multi-objective problem. Experiments show that our DeCoNASNet outperforms the state-of-the-art lightweight super-resolution networks designed by handcraft methods and existing NAS-based design.
\end{abstract}

% no keywords

% For peer review papers, you can put extra information on the cover
% page as needed:
% \ifCLASSOPTIONpeerreview
% \begin{center} \bfseries EDICS Category: 3-BBND \end{center}
% \fi
%
% For peerreview papers, this IEEEtran command inserts a page break and
% creates the second title. It will be ignored for other modes.
\IEEEpeerreviewmaketitle

\section{Introduction}
% no \IEEEPARstart
Single image super-resolution (SISR) is a task that creates a clearer high-resolution image from a single low-resolution input. It is an important technology that can be used as a pre-processing step to increase performances of various tasks such as medical image analysis~\cite{isaac2015super}, satellite image recognition~\cite{luo2017video}, security image processing~\cite{zou2011very}, etc. The SISR is an ill-posed problem because multiple HR images can be mapped to a single LR image. Hence, learning-based methods trained with many LR-HR image pairs are generally more effective than the interpolation-based \cite{zhang2006edge} or reconstruction-based methods~\cite{zhang2012single}.
 
Recently, many deep neural networks for SISR have been developed~\cite{dong2014learning, dong2016accelerating, shi2016real, kim2016accurate, lim2017enhanced, tong2017image, tai2017memnet, zhang2018residual,ahn2018fast, lai2017deep, li2018multi}, where Dong \MakeLowercase{\textit{et al.}}'s SRCNN \cite{dong2014learning} is the first convolutional neural network (CNN) for the SISR. It consists of three convolution layers and yet outperformed conventional non-learning methods by a large margin. FSRCNN~\cite{dong2016accelerating} and ESPCN~\cite{shi2016real} tried to reduce the computational cost of the structure. They used LR images directly as the input of their neural networks. Then, deconvolution and sub-pixel convolution layers were used for upsampling their results. VDSR~\cite{kim2016accurate} dramatically increased the depth of the model by residual learning and gradient clipping strategy. Lim \MakeLowercase{\textit{et al.}}~\cite{lim2017enhanced} further improved performance by a residual block composed of extensive features (EDSR) and multi-scale structure (MDSR). In MemNet~\cite{tai2017memnet}, MSRN~\cite{li2018multi}, and DenseSR~\cite{tong2017image}, they proposed specific blocks in their models, such as memory block, multi-scale residual block or dense block. SelNet~\cite{choi2017deep} used selection unit instead of conventional Relu operation. Zhang {\em et al.} proposed RDN~\cite{zhang2018residual}, which consists of residual dense block and dense feature fusion that extract abundant information from the input. RCAN~\cite{zhang2018image} applied channel attention mechanism to improve representational ability of CNNs. It is believed that most of these networks have been designed through laborious trials of human experts, by tuning a large number of network hyperparameters such as operation type, the number of channels, connection, depth, etc. However, a network designed by human labor may not be optimal for the given resources.
 
\begin{figure*}[t]
	\centering
	\includegraphics[width=0.95\textwidth]{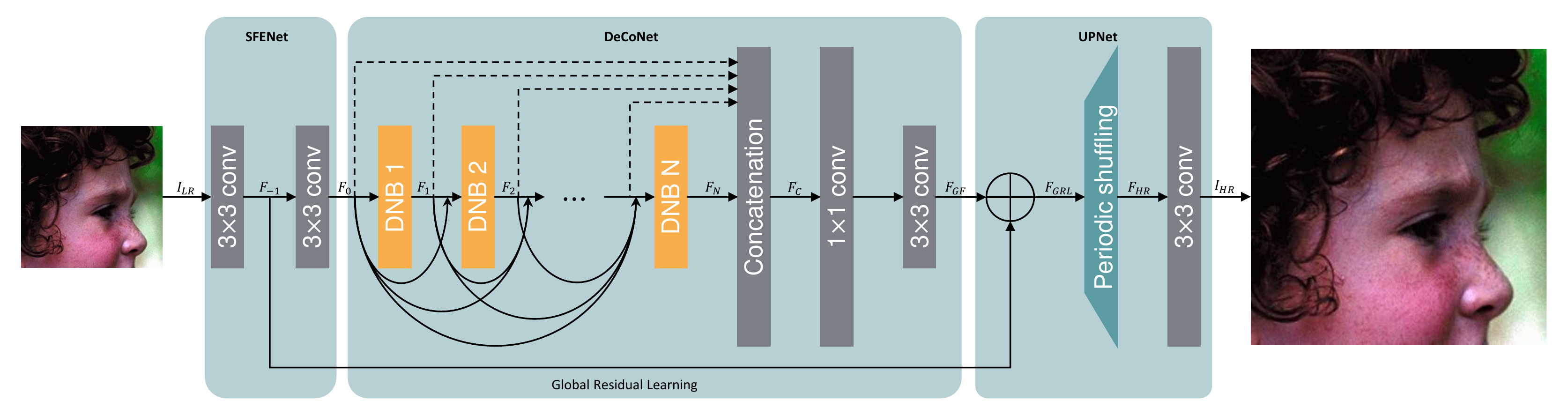}
	\caption{The architecture of the proposed DeCoNASNet system.}
	\label{deco_structure}
\end{figure*}

In the case of image classification research fields, neural architecture search (NAS) methods have been proposed \cite{zoph2016neural, liu2018progressive, pham2018efficient, xie2017genetic, lu2019nsga, real2019regularized, liu2018darts, luo2018neural}, which automatically find an optimal network to alleviate human labors \cite{zoph2016neural, liu2018progressive, pham2018efficient, xie2017genetic, lu2019nsga, real2019regularized, liu2018darts, luo2018neural}.
As a pioneering study, Zoph \MakeLowercase{\textit{et al.}}~\cite{zoph2016neural} proposed a controller network that generates a child network structure based on reinforcement learning (RL). The NAS trained the controller network by REINFORCE~\cite{williams1992simple}, which is a kind of policy gradient algorithm. But, this method took a tremendous amount of time to evaluate the candidate models because they trained the models from scratch. To reduce the time it takes to measure the accuracy, Liu \MakeLowercase{\textit{et al.}}~\cite{liu2018progressive} proposed the PNAS, which used the sequential model-based optimization (SMBO) and learned a surrogate model to predict its performance directly. Also, ENAS~\cite{pham2018efficient} reduced the evaluation time by about a thousand times, by applying a weight sharing scheme. The ENAS constructed a large graph and regarded each model as a sub-graph of the main graph. In this way, child networks can share their parameters while being trained separately.

Another branch of NAS algorithms is the evolutionary-based methods \cite{xie2017genetic, lu2019nsga, real2019regularized}, which pick a population of neural networks randomly. Then, they encode the network structures as binary sequences and apply genetic modifications such as mutation and crossover to find better models. Additionally, NSGA-Net~\cite{lu2019nsga} used Bayesian optimization to get an advantage from its search history. Real \MakeLowercase{\textit{et al.}}~\cite{real2019regularized} introduced Amoeba-Net, which use an aging evolution algorithm to discard the earliest trained network. DARTS~\cite{liu2018darts} and NAO~\cite{luo2018neural} are also promising architectures that are approached differently from RL and evolutionary-based algorithm. DARTS optimized all parameters and connections in the neural architecture jointly with a continuous relaxation of the search space. NAO proposed a learnable embedding space of architectures and found the best model from it.

Recently, some researchers expanded the NAS to other domains such as object detection~\cite{liu2019auto} (Auto-deeplab) and image SR~\cite{chu2019multi} (MoreMNAS), \cite{chu2019fast} (FALSR). Specifically, they used a reinforced evolution algorithm and solved the image SR task as a multi-objective problem. However, these reinforced evolution algorithms need more than 1 GPU month to find optimal architectures. Furthermore, FALSR evaluated the network approximately because they did not use a complete training strategy.

To overcome these drawbacks, we take the idea of ENAS~\cite{pham2018efficient} as our baseline search framework. The ENAS consists of a controller and child networks, where the controller is composed of LSTM blocks to generate a child network sequence. We also use the REINFORCE~\cite{williams1992simple} algorithm to train the controller in the direction of increasing PSNR of the SR results. As in the original ENAS for the classification problems, child networks share their parameters during the training and evaluation. In addition, we propose a complexity-based penalty to reduce the rewards for the networks that need a large number of parameters. We exclude redundant hierarchical information by predicting connections for local and global feature fusion through the controller.

SR is a type of regression that generally needs a deeper and more complex network than classification. Hence, in this paper, we search for a new SR architecture on the densely constructed search space. Specifically, we propose a Densely Connected Neural Architecture Search (DeCoNAS) method, which attempts to find optimal connections on the baseline of residual dense network (RDN)~\cite{zhang2018residual}. The proposed DeCoNAS search space consists of mix nodes for densely connected network blocks (DNB), local feature fusion, and global feature fusion. In addition to the convolution layer with $3 \times 3$ filters in the baseline network, dilated convolution~\cite{yu2015multi} and depth separable convolution~\cite{chollet2017xception} are included in the search space for better performance. Experiments show that our DeCoNASNet performs better than human-crafted networks and the existing NAS-based SR network \cite{chu2019multi,chu2019fast}

Our main contributions are summarized as follows:

\subsubsection{A New NAS-Based SR}
We propose a new NAS-based SR network design, named DeCoNAS, which searches for networks with higher performance by combining hierarchical and local information efficiently.

\subsubsection{Complexity-Based Penalty}
We design a complexity-based penalty and add it to the reward of the REINFORCE algorithm, which enables us to search for an efficient network that has high performance and fewer parameters.

\subsubsection{Feature Fusion Layer Search}
We also search for efficient feature fusion method. Instead of connecting all the global/local feature fusion layers, we design the connection to be predicted through the controller, which removes redundant hierarchical information and hence reduce network complexity.

\begin{figure}[t]
	\centering
	\includegraphics[width=0.9\columnwidth]{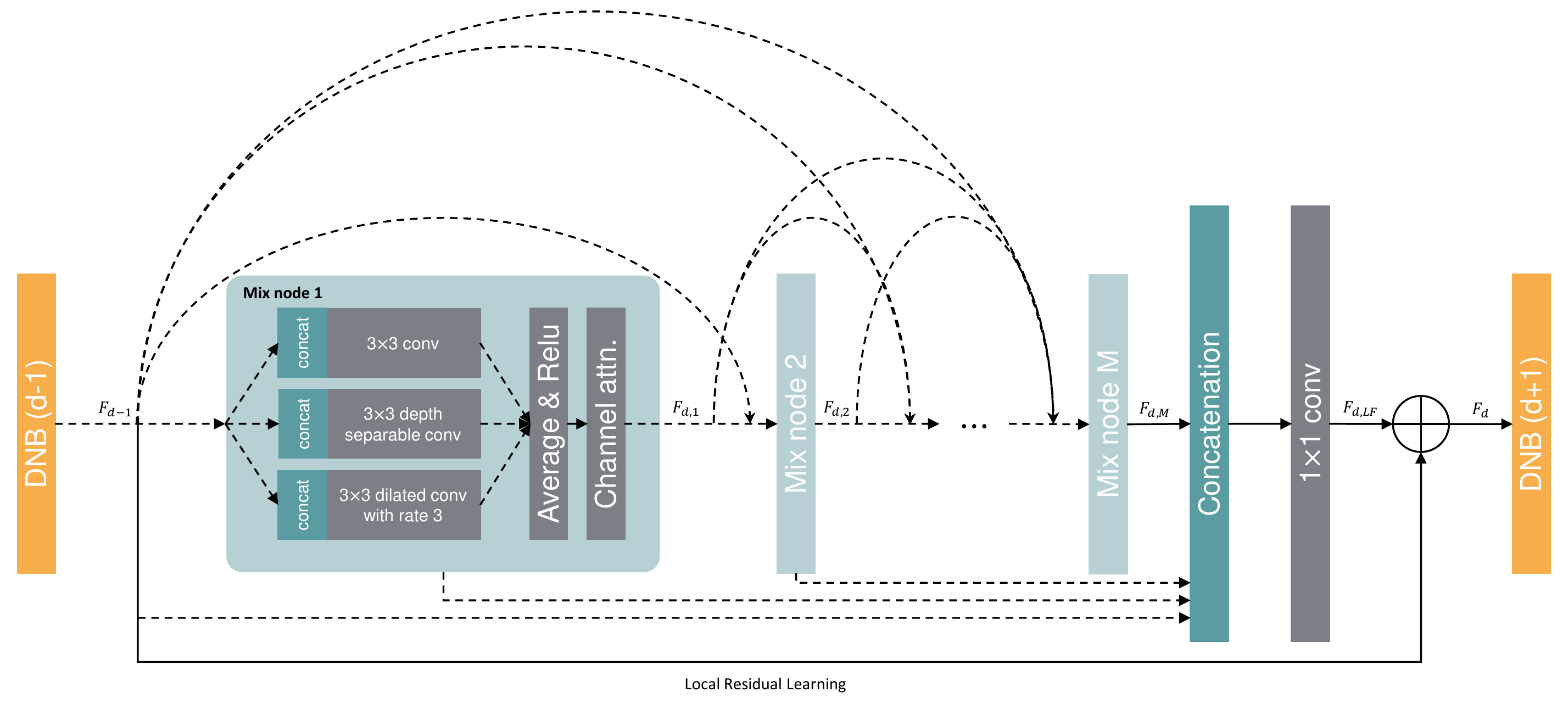}
	\caption{Detailed structure of the densely connected network blocks (DNB).}
	\label{dnb_structure}
\end{figure}

\section{Proposed Method}
As a typical RL framework, the proposed architecture consists of two parts: a child network (denoted as DeCoNASNet) for reward measurement and a controller for network structure generation. Following ENAS~\cite{pham2018efficient}, we try to save time by using parameter sharing when training a child network. Also, we regard the SR as a multi-objective task. That is, we design a complexity-based penalty to consider not only the PSNR but also the parameter complexity of the network for calculating the reward.

\begin{figure*}[t]
\centering
\subfloat[]{
	\includegraphics[width=0.55\linewidth]{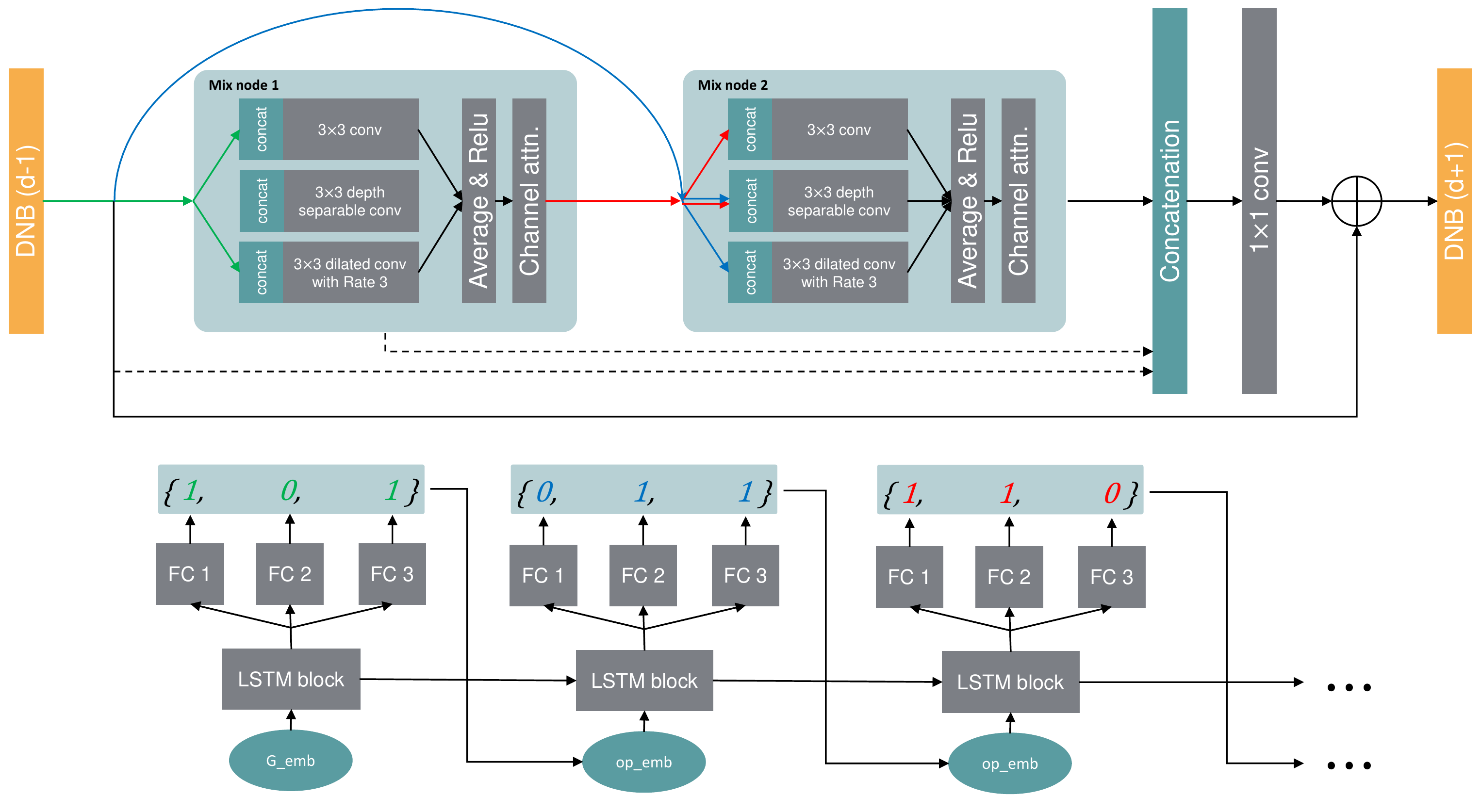}
} 
\qquad
\subfloat[]{
	\includegraphics[width=0.34\linewidth]{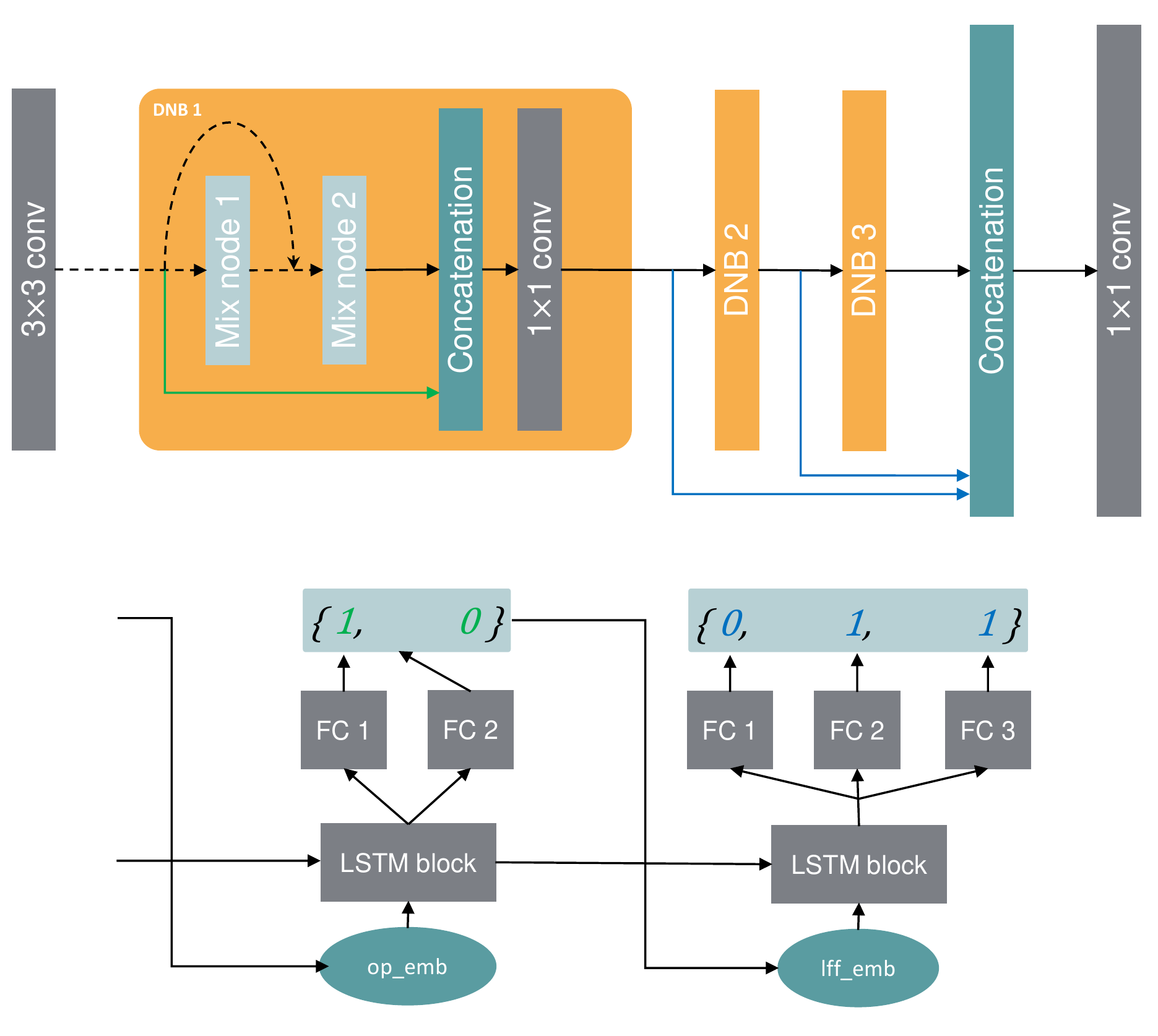}
}
\caption{The example structure of our Controller and DeCoNASNet structure with 2 mix nodes ($M=2$) and 3 DNBs ($N=3$). (a) is the example for mix nodes, and (b) is the example for feature fusion layers.}
\label{deco_example}
\end{figure*}

We use a two-layer LSTM network in our controller to generate the DeCoNASNet structure. Supposing that the child network $\textbf{c}$ consists of $N$ blocks, $M$ mix nodes, and $K$ mix node operations, the controller sequence $\textbf{S}_\textbf{c}$ for $\textbf{c}$ is
\begin{equation}
\begin{split}
\textbf{S}_\textbf{c} & = \{ \textbf{S}_M, \textbf{S}_F \}, \\
\textbf{S}_M & = \{ S_{i,j}^k  \},  0 < i \leq M, 0 \leq j < i, 0 \leq k < K, \\
\textbf{S}_F & = \{ {S_l}^0:{S_l}^{M-1}, {S_g}^0:{S_g}^{N-1} \},
\end{split}
\label{eq_controller_seq}
\end{equation}
where $\textbf{S}_M$ is the sequence for mix node configuration, and $\textbf{S}_F$ is the sequence for the feature fusion layer. $\textbf{S}_F$ consists of two sequences, $\textbf{S}_l$ and $\textbf{S}_g$, which denote the sequence for local feature fusion and global feature fusion, respectively.

\subsection{Overall structure of DeCoNASNet}
DeCoNASNet consists of three parts as shown in Fig.~\ref{deco_structure}, inspired by the RDN~\cite{zhang2018residual} architecture: shallow feature extractor network (SFENet), densely connected network (DeCoNet), and UPNet. The output of SFENet can be represented as
\begin{equation}
F_0=H_{SFE2}(H_{SFE1}(I_{LR})),
\end{equation}
where $H(\cdot)$ denotes the convolution operation. SFENet converts the input image $I_{LR}$ into a shallow feature $F_0$, which is used as the input to the DeCoNet. Then, the output of the $d$-th DNB in DeCoNet, denoted by $F_d$ is expressed as
\begin{equation}
F_d=H_{DNB,d}(H_{1}(concat(F_0, F_1, \dots, F_{d-1}))),
\end{equation}
where $H_{DNB}(\cdot)$ denotes the DNB operation and $H_{1}(\cdot)$ is $1 \times 1$ convolution to match the input channels of DNBs. We will explain the operation $H_{DNB}(\cdot)$ in the next subsection. We omit $H_{1}(\cdot)$ and concatenation operation in Fig.~\ref{deco_structure} for simplicity. The global feature fusion layer follows after the $N$-th DNB to combine the information of the features of DNB. The output of the global feature fusion layer, denoted by $F_{GF}$ is described as
\begin{equation}
\begin{split}
F_{GF} & = H_{GFF}(F_C), \\
F_C & = concat({S_g}^0 \cdot F_0, \dots, {S_g}^{N-1} \cdot F_{N-1}, F_N),
\end{split}
\label{eq_gff}
\end{equation}
where $H_{GFF}(\cdot)$ denotes the $1 \times 1$ convolution and $3 \times 3$ convolution operation, ${S_g}^i$ denotes the output global feature fusion sequence of controller, and $concat(\cdot)$ denotes the concatenation of features. We omit $F_i$ when concatenating features if ${S_g}^i$ is zero. All of the ${S_g}^i$ equal to one if we do not use the feature fusion search strategy. The UPNet combines the output of DeCoNet and $F_{-1}$, which is the shallow feature from the SFENet. We use periodic shuffling operation and applied $3 \times 3$ convolution as in ESPCN~\cite{shi2016real}, to convert LR features to high-resolution images. We fix the SFENet and the UPNet,  while we search for the DeCoNet. In all of our figures, we use dashed arrows to depict that the connection is to be searched.

\subsection{Constructing the DNB}
We apply the same operation $H_{DNB}(\cdot)$ through all the DNBs. Each DNB consists of M mix nodes as shown in Fig.~\ref{dnb_structure}, where there are three element candidates in the mix node:
\begin{itemize}
\item $3 \times 3$ 2D convolution,
\item $3 \times 3$ depth separable convolution,
\item $3 \times 3$ dilated convolution with rate $3$.
\end{itemize}
Also, $F_{d,m}$ over the arrow is the output of the $m$-th mix node in the $d$-th DNB with K candidate mix node operations, which are obtained as
\begin{equation}
F_{d,m} = 
\begin{cases}
F_{d,m-1}, & \mbox{if } \textbf{S}_{m,0:m-1}^{0:K-1} = \textbf{0} \\
CA(Relu(average(\textbf{F}_{inter}))), & \mbox{else }
\end{cases}
\label{eq_fdm}
\end{equation}
where
\begin{equation}
\begin{split}
\textbf{F}_{inter} & = {F_{d,m}^{0:K-1}}, \\
F_{d,m}^i & = H_i(concat(S_{m,0}^i \cdot F_{d,0}, \dots, S_{m,m-1}^i \cdot F_{d,m-1}))
\end{split}
\label{eq_mix_node_feature}
\end{equation}
where $H_i(\cdot)$ denotes the $i$-th operation in $K$ operations, and $\textbf{S}_{m,0:m-1}^{0:K-1}$ is the sequence for the $m$-th mix node configuration. Same as the Eq.~(\ref{eq_gff}), we omit $F_{d,m}^i$ or $F_{d,m}$ if $\textbf{S}_{m,0:m-1}^i = \textbf{0}$ or $S_{m,j}^i = 0$ in Eq.~(\ref{eq_mix_node_feature}). $CA(\cdot)$ in Eq.~(\ref{eq_fdm}) denotes channel attention network used in RCAN~\cite{zhang2018image}.

\subsection{Constructing controller for the DeCoNASNet}

\subsubsection{Controller output of the mix node}
We need $i \times K$ sequences to create the $i$-th mix node. Hence, our controller is composed of $\sum_{i=1}^{M}i = \frac{M(M+1)}{2}$ LSTM blocks, and $K$ fully connected layers are connected to each LSTM block. For example, As shown in Fig.~\ref{deco_example}(a), we use 3 LSTM blocks and 9 outputs to create the connections for two mix nodes.

\subsubsection{Controller output of the feature fusion layer}
There are two LSTM blocks for feature fusion layer search. These blocks are connected to the last LSTM block for mix node, in order to include the information about the mix node structure. Like the mix node, we connect $N$ and $M$ fully connected layers to two LSTM blocks, respectively. The output from each LSTM block denotes the connection between the mix node and the feature fusion layer. Fig.~\ref{deco_example}(b) shows an example connection of local/global feature fusion layer.

\subsection{Training DeCoNAS and complexity-based penalty}	
Following ENAS~\cite{pham2018efficient}, the DeCoNAS has two learnable parameters. The parameter of the controller is $\boldsymbol{\theta}$, and the parameter of the child network is $\textbf{w}$. To learn $\boldsymbol{\theta}$ and $\textbf{w}$ alternately, we use a two-step learning strategy. In the first step, we train $\textbf{w}$ using training data. We use an RL scheme to train $\boldsymbol{\theta}$, with the reward signal consisting of peak signal to noise ratio (PSNR) and complexity-based penalty.

\subsubsection{Training child network}
As the first step to training DeCoNAS, we need to learn $\textbf{w}$, which is the parameter of the child network, $\textbf{c}$. This problem is defined as
\begin{equation}
 \min_\textbf{w} \E_{\textbf{c}\sim\pi(\textbf{c};\boldsymbol{\theta})}[L(\textbf{c};\textbf{w})].
 \label{eq_child_opt}
\end{equation}

To optimize $\textbf{w}$, we fix the controller's policy $\pi(\textbf{c};\boldsymbol{\theta})$ and use Adam optimizer~\cite{kingma2015adam}. In this case, we use the L1 loss for $L(\textbf{c}; \textbf{w})$, calculated on training data and model $\textbf{c}$ generated by $\pi(\textbf{c};\boldsymbol{\theta})$. Gradient of $\E_{\textbf{c}\sim\pi(\textbf{c};\boldsymbol{\theta})}[L(\textbf{c};\textbf{w})]$ is calculated by Monte Carlo estimate
\begin{equation}
\bigtriangledown_\textbf{w} \E_{\textbf{c}\sim\pi(\textbf{c};\boldsymbol{\theta})}[L(\textbf{c};\textbf{w})] \approx \frac{1}{M}\sum_{i=1}^{M}\bigtriangledown_\textbf{w} L(\textbf{c}_i;\textbf{w}),
\label{eq_monte_carlo}
\end{equation}
where $\textbf{c}_i$'s are sampled by the controller's policy $\pi(\textbf{c};\boldsymbol{\theta})$.
As mentioned in ENAS~\cite{pham2018efficient}, $\textbf{w}$ can be optimized by calculating the gradient for only one model $\textbf{c}$ generated by $\pi(\textbf{c};\boldsymbol{\theta})$ for each mini-batch.

\subsubsection{Training controller with performance reward and complexity-based penalty}
In the second step, we need to train $\boldsymbol{\theta}$, which is the controller's parameter. This problem is to maximize the expected reward as 
\begin{equation}
\max_{\boldsymbol{\theta}} E_{P(a_{1:T};\boldsymbol{\theta})}[R],
\label{eq_controller_opt}
\end{equation}
where $a_{1:T}$ is the controller output for the child network $\textbf{c}$, which follows the distribution of $\pi(\textbf{c};\boldsymbol{\theta})$. We compute the gradient of the problem by using the approximation of gradients in REINFORCE~\cite{williams1992simple} as
\begin{equation}
\bigtriangledown_{\boldsymbol{\theta}} \E_{P(a_{1:T};\boldsymbol{\theta})}[R] = \sum_{t=1}^{T}[\bigtriangledown_{\boldsymbol{\theta}} \log P(a_t|a_{t-1:1};\boldsymbol{\theta})(R-b)]
\label{eq_reinforce}
\end{equation}
where $b$ is the baseline for reducing the variance, which is the moving average of the reward.
While ENAS used classification accuracy for $R$, we calculate it  differently as
\begin{equation}
R = p(\textbf{c},\textbf{w}) – \alpha*cb(\textbf{c}),
\label{eq_reward}
\end{equation}
where
$p(\textbf{c},\textbf{w})$ is the PSNR of model $\textbf{c}$. We calculate the PSNR using the validation set rather than the training set to prevent overfitting. Also, $cb(\textbf{c})$ is the complexity-based penalty, calculated as 
\begin{equation}
cb(\textbf{c}) = \frac{n_m}{n_{cm}},
\label{eq_complexity_penalty}
\end{equation}
where $n_m$ denotes the number of parameters in the generated model $\textbf{c}$ and $n_{cm}$ indicates the number of parameters in the most complex model in the search space. We also multiply $\alpha$ by $cb(\textbf{c})$, allowing the user to set a trade-off between the performance and model complexity. Finally, we use Adam optimizer~\cite{kingma2015adam} to maximize the reward.

\section{Experimental results}

\subsection{Settings}
\subsubsection{Datasets and metrics}
We use DIV2K dataset \cite{timofte2017ntire} for the training, which has been widely used for training image restoration networks. The DIV2K contains 800 training images, 100 validation images, and 100 test images. We use all images in the training set to train our DeCoNASNet, and use all of the validation images when calculating the reward and training the controller. Experiments are conducted on four benchmark datasets, Set5~\cite{bevilacqua2012low}, Set14~\cite{zeyde2010single}, B100~\cite{martin2001database}, and Urban100~\cite{huang2015single}, where we compute PSNR and SSIM~\cite{wang2004image} on the Y channel.

\subsubsection{Implemenation details}
Our proposed DeCoNASNet has 4 DNBs, and each DNB has 4 mix nodes. The output channel of SFENet and DNB are both 64. The $3 \times 3$ convolution layer in UPNet also has 64 output channels, and we conduct periodic shuffling on the feature maps. The final convolution layer has $3\times3$ filters and three output channels to restore the high-resolution images.

The LSTM block in the controller is made of two stacked LSTM layers with 64 hidden states. Fully connected layers for one operation in each LSTM block are sharing their parameters. We tie the LSTM outputs with word embeddings~\cite{inan2016tying} to make the input of the next LSTM block.

\subsubsection{Training setting}
In the search phase, we need to train controller and DeCoNASNet together. We use variance scaled initialization~\cite{he2015delving} with 0.02 scaling value for DeCoNASNet parameter $\textbf{w}$ and controller parameter $\boldsymbol{\theta}$. To train the controller, we apply 100 iterations for one epoch, and the learning rate is fixed to $3\times10^{-4}$.
\begin{table}[t]
	\begin{center}
		\caption{Performance comparison between search settings.}
		\label{tab_perf_4models}
		\begin{tabular}{c|c|c|c|c|c} % <-- Alignments: 1st column left, 2nd middle and 3rd right, with vertical lines in between
			\toprule[2 pt]
			\multicolumn{2}{c|}{Search setting} & CB0\_FF0 & CB0\_FF1 & CB1\_FF0 & CB1\_FF1 \\
			\midrule
			\multirow{2}{*}{Best}& PSNR & \textbf{35.919} & 35.602 & 35.894 & 35.436 \\			 
			& Parameters & 22.9 M & 25.4 M & \textbf{18.0 M} & 25.9 M \\
			\midrule
			\multirow{2}{*}{Mean}& PSNR & 35.294 & 35.168 & \textbf{35.324} & 35.025 \\			 
			& Parameters & 22.5 M & 28.4 M & \textbf{17.1 M} & 24.9M \\
			
			\bottomrule[2 pt]
		\end{tabular}
	\end{center}
\end{table}
\begin{figure}[t]
	\centering
	\includegraphics[width=0.9\columnwidth]{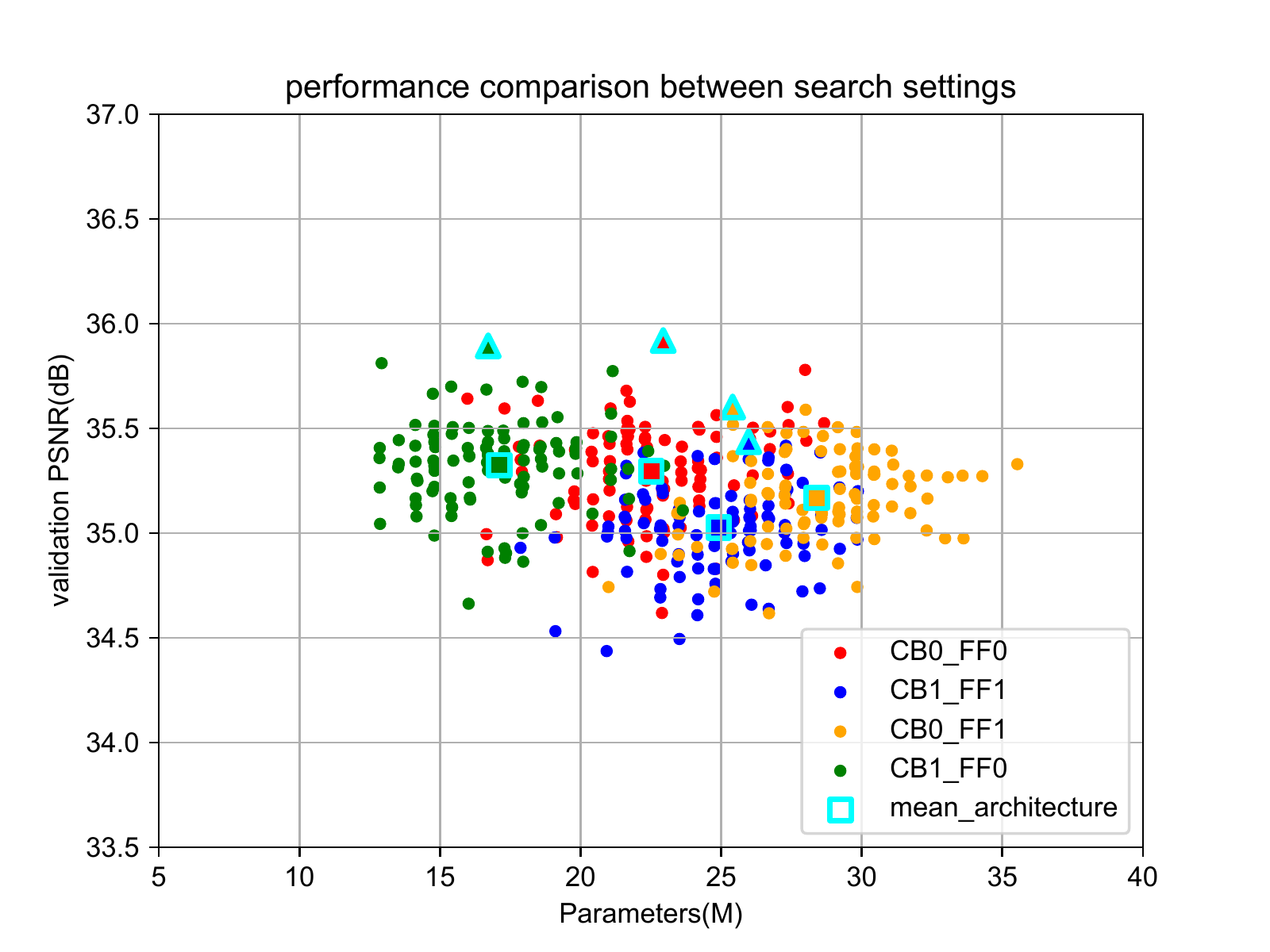}
	\caption{Scatter plot of 100 samples for each search setting.}
	\label{fig_perf_4models}
\end{figure}
For training the DeCoNASNet, we randomly extract 16 LR patches of size $64 \times 64$ from the DIV2K training image as the input to the network. After extracting the patches, we apply horizontal flip and 90\textdegree, 180\textdegree, 270\textdegree  ~rotations to each patch randomly for data augmentation. We conduct 1,000 backpropagation for an epoch, where Adam optimizer~\cite{kingma2015adam} is used for updating parameters. The learning rate is initialized to $10^{-4}$ and decreased by half for every $5\times10^5$ iterations (50 epochs), and 200 epochs are conducted for the search phase. We sample 100 candidate structures by the trained controller and choose the best architecture as our DeCoNASNet structure. After choosing the best architectures, we train the DeCoNASNet for 1,000 epochs. The learning rate is initialized to $10^{-4}$ and decreases by half for every 200 epochs. The other settings are the same as the search phase.

\begin{figure}[t]
	\centering
	\includegraphics[width=0.8\columnwidth]{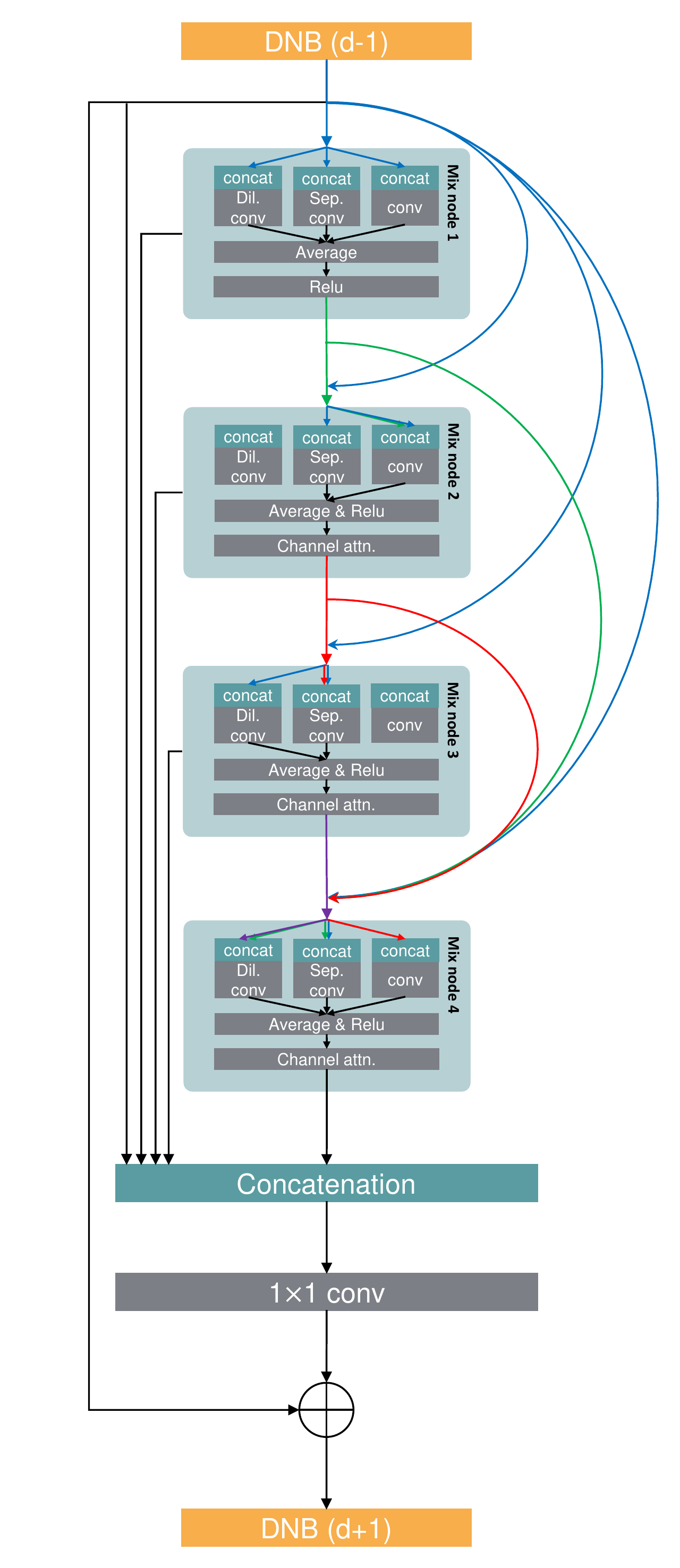}
	\caption{Our DeCoNASNet model found by controller.}
	\label{fig_DeCoNASNet_l}
\end{figure} 

\begin{table*}[t]
	\begin{center}
		\caption{Public benchmark test results (PSNR/SSIM) for $\times 2$ SR. The red color means the best performance and the blue means the second best. The ``Design time'' at the last column indicates the times taken by the NAS approaches.}
		\label{tab_perf_sota}
		\begin{tabular}{l|c|c|c|c|c|c} % <-- Alignments: 1st column left, 2nd middle and 3rd right, with vertical lines in between
			\toprule[2 pt]
			Model & Params & Set 5 & Set 14 & B100 & Uban100 & Design time\\
			Bicubic &  - & 33.66 / 0.9299 & 30.24 / 0.8688 & 29.56 / 0.8431 & 26.88 / 0.8403 & - \\
			SRCNN~\cite{dong2014learning} &  57K & 36.66 / 0.9542 & 32.45 / 0.9067 & 31.36 / 0.8879 & 29.50 / 0.8946 & - \\
			VDSR~\cite{kim2016accurate} &  665K & 37.53 / 0.9587 & 33.03 / 0.9124 & 31.90 / 0.8960 & 30.76 / 0.9140 & - \\
			LapSRN~\cite{lai2017deep} &  813K & 37.52 / 0.9591 & 33.08 / 0.9130 & 31.80 / 0.8950 & 30.41 / 0.9101 & - \\
			MemNet~\cite{tai2017memnet} &  677K & 37.78 / \color{blue}0.9597 & 33.28 / 0.9142 & 32.08 / 0.8978 & 31.31 / 0.9195 & - \\
			SelNet~\cite{choi2017deep} &  970K & \color{blue}37.89 \color{black}/ \color{red}0.9598 & \color{blue}33.61 \color{black}/ 0.9160 & 32.08 / 0.8984 & - / - & - \\
			CARN~\cite{ahn2018fast} &  1,582K & 37.76 / 0.9590 & 33.52 / 0.9166 & 32.09 / 0.8978 & 31.92 / \color{blue}0.9256 & - \\
			MoreMNAS-A~\cite{chu2019multi} &  1,039K & 37.63 / 0.9584 & 33.23 / 0.9138 & 31.95 / 0.8961 & 31.24 / 0.9187 & 56 GPU days \\
			FALSR-A~\cite{chu2019fast} &  1,021K & 37.82 / 0.9595 & 33.55 / \color{blue}0.9168& \color{blue}32.12 \color{black}/ \color{red}0.8987 & \color{blue}31.93 \color{black}/ \color{blue}0.9256 & 24 GPU days\\
			DeCoNASNet (ours) &  1,713K & \color{red}37.96 \color{black}/ 0.9594 & \color{red}33.63 \color{black}/ \color{red}0.9175 & \color{red}32.15 \color{black}/ \color{blue}0.8986 & \color{red}32.03 \color{black}/ \color{red}0.9265 & 12 GPU hours\\
			\bottomrule[2 pt]
		\end{tabular}
	\end{center}
\end{table*}

\begin{figure*}[t]
	\centering
	\includegraphics[width=0.9\textwidth]{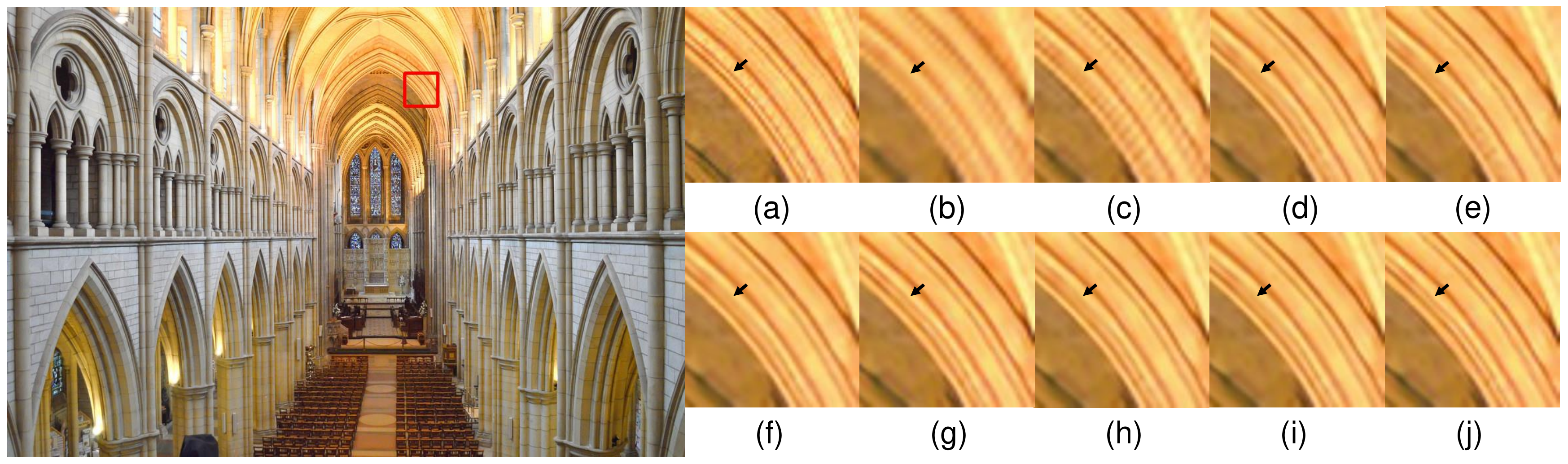}
	\caption{Qualitative comparison of the conventional methods and ours. (a) HR image, (b) bicubic LR image, (c) SRCNN~\cite{dong2014learning}, (d) VDSR~\cite{kim2016accurate}, (e) LapSRN~\cite{lai2017deep}, (f) MemNet~\cite{tai2017memnet}, (g) CARN~\cite{ahn2018fast}, (h) MoreMNAS~\cite{chu2019multi}, (i) FALSR~\cite{chu2019fast}, (j) DeCoNASNet (ours).}
	\label{qualitative_results}
\end{figure*}

\subsection{Results}
\subsubsection{DeCoNAS search result}
We use 16 DNBs ($N=16$) and 8 mix nodes in each DNB ($M=8$) to verify the effect of feature fusion strategy and complexity-based penalty. Table~\ref{tab_perf_4models} and Fig.~\ref{fig_perf_4models} show the performance of total 400 DeCoNAS structures in four settings (100 structures each). The baseline setting (denoted as FF0\_CB0), which omitted the feature fusion search (FF) and complexity-based penalty (CB), shows the best performance with 23.2 M parameters. We add one of CB or FF to FF0\_CB1 and FF1\_CB0. From the results, we can also see that FF0\_CB1 achieves almost the same performance as the baseline search strategy, with 20\% less parameters, and FF1\_CB0 has slightly lower performance with large parameters. Finally, we apply both strategies, resulting in FF1\_CB1, which is shown to have fewer parameters than the FF1\_CB0 model, but yields inferior performance than the others. Further discussions about CB and FF strategies are at the ablation study section.

We use 4 DNBs ($N=4$) and 4 mix nodes ($M=4$) in each DNB  to make DeCoNASNet structure. We choose the best architecture which belongs to FF0\_CB1 setting. Fig.~\ref{fig_DeCoNASNet_l} shows the DNB structure of DeCoNASNet:\{7, 6, 4, 3, 0, 2, 2, 3, 4, 1\}. The local/global feature fusion connections are all connected because we do not use the feature fusion search strategy. We regard three outputs of each controller LSTM block as the binary number and convert it to a decimal number for simplicity. For example, four in the sequence means \{1, 0, 0\} and six is \{1, 1, 0\}. It takes about 12 hours by 1 Titan XP GPU to search for the DeCoNASNet structure, which is far less than other NAS-based methods.

\subsubsection{Comparison with state-of-the-art methods}
We compare our DeCoNASNet with six lightweight networks (SRCNN~\cite{dong2014learning}, VDSR~\cite{kim2016accurate}, MemNet~\cite{tai2017memnet}, LapSRN~\cite{lai2017deep}, SelNet~\cite{choi2017deep}, CARN~\cite{ahn2018fast}) and two NAS-based methods (MoreMNAS~\cite{chu2019multi}, FALSR~\cite{chu2019fast}). The results are shown in Table~\ref{tab_perf_sota}, where we can see that DeCoNASNet outperforms other hand-crafted lightweight models and existing NAS-based ones while using somewhat more parameters, still within 2M. The most important advantage of our method is that it finds the optimal structure within 16 hours, which is $\times50$ faster than other NAS-based methods.

We compare the visual results of our model with the others (SRCNNN~\cite{dong2014learning}, VDSR~\cite{kim2016accurate}, LapSRN~\cite{lai2017deep}, MemNet~\cite{tai2017memnet}, CARN~\cite{ahn2018fast}, MoreMNAS~\cite{chu2019multi}, FALSR~\cite{chu2019fast}) in Fig.~\ref{qualitative_results}. We find that DeCoNASNet successfully restores the details in images. Specifically, DeCoNASNet and CARN restore the double curves at the black arrow, while others do not.

\begin{figure}[t]
	\centering
	\includegraphics[width=0.9\columnwidth]{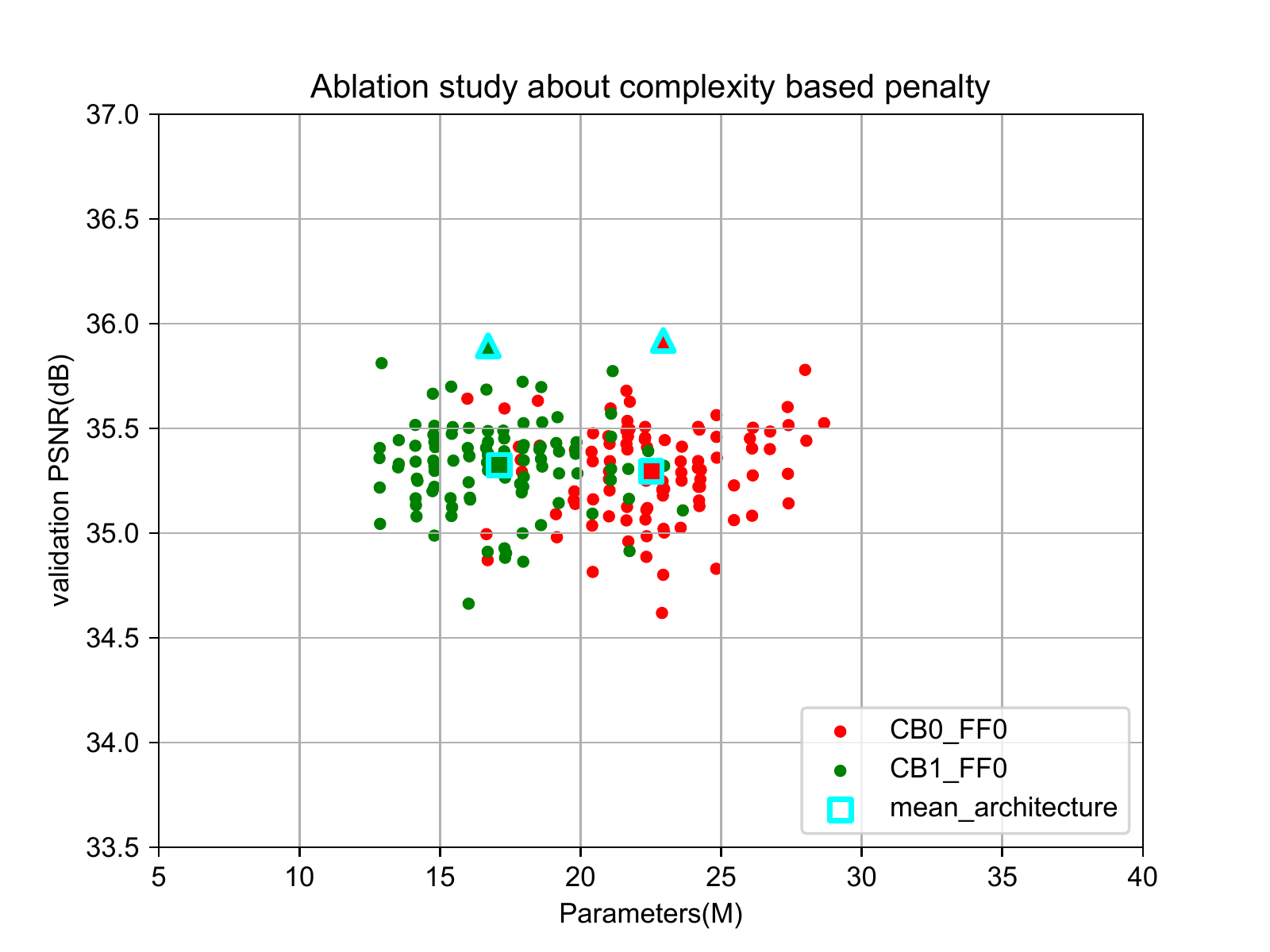}
	\caption{Ablation analysis about effect of the complexity-based penalty.}
	\label{fig_ablation_cb}
\end{figure} 
\begin{figure}[t]
	\centering
	\includegraphics[width=0.9\columnwidth]{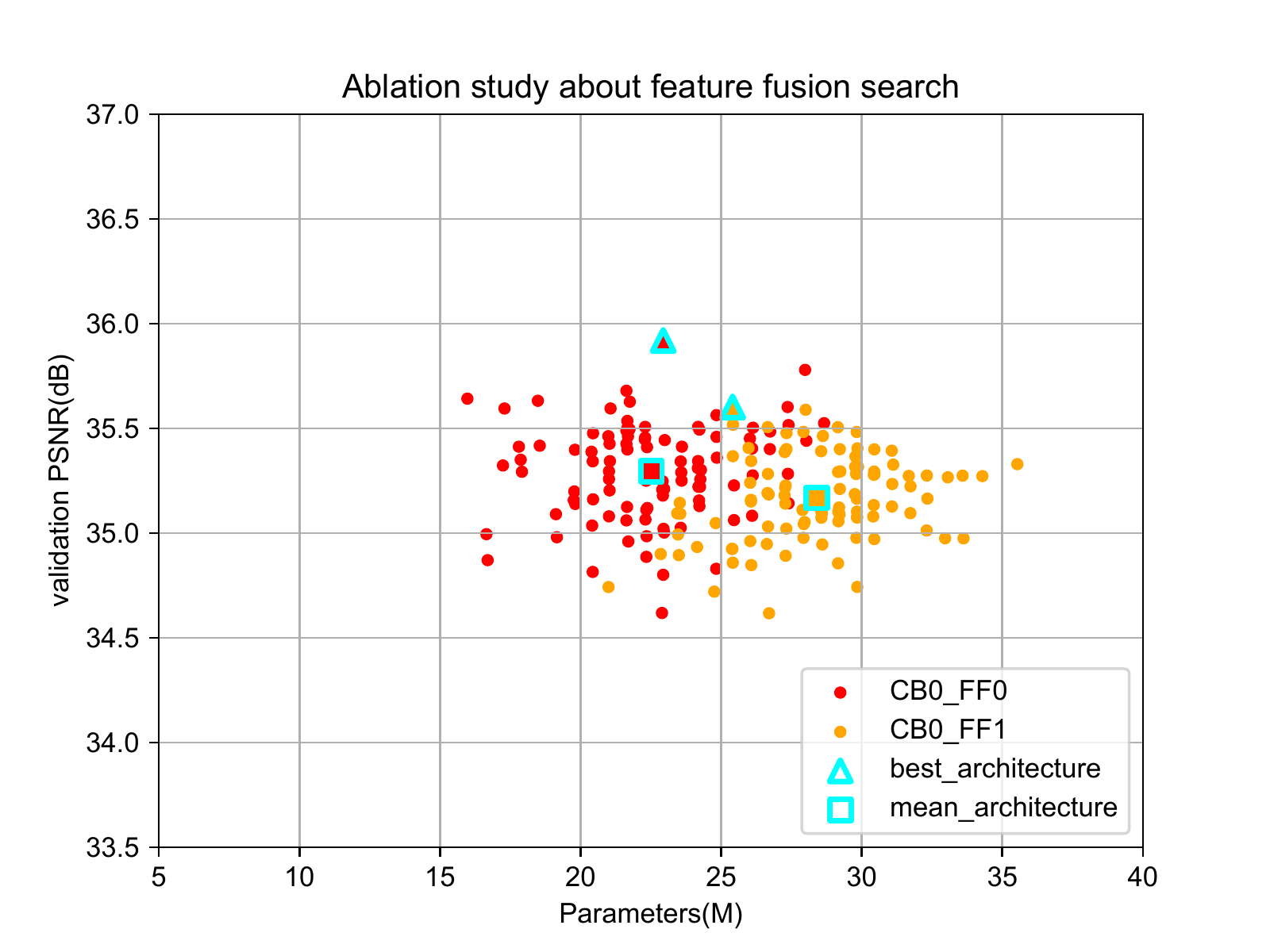}
	\caption{Ablation analysis about the feature fusion search strategy.}
	\label{fig_ablation_ff}
\end{figure} 

\subsection{Ablation study}
Fig.~\ref{fig_ablation_cb} shows a comparison between different complexity-based penalty coefficients, $\alpha$. We conduct two experiments in the DeCoNAS search space ($N=16$, $M=8$) to analyze the effect of the complexity-based penalty. The baseline strategy is denoted as CB0\_FF0, and the other strategy using CB coefficient $\alpha=2$ is described as CB1\_FF0. The parameters of the searched model are decreased, but the performance difference is small when we change the CB strategy. Hence, we can validate that the complexity-based penalty efficiently controls the trade-off between the performance and the number of parameters. Users can choose $\alpha$ to find models that fit their purpose.

We also show the effect of feature fusion search (FF) strategy in Fig.~\ref{fig_ablation_ff}. The red and orange dots in the figure denote CB0\_FF0 and CB0\_FF1, respectively. It can be seen that the FF strategy tends to find more complex models than no FF strategy. This is mainly because the controller tries to compensate for the information loss from the feature fusion layer disconnection. We can verify that the connections to the global/local feature fusion layer are more important than the connections in the mix node.

\section{Conclusion}
We have proposed an RL-based neural architecture search algorithm for image SR, named as DeCoNAS. It is shown that the proposed method can find a promising lightweight SR network (DeCoNASNet) within 16 hours, which is a lot faster than other NAS-based algorithms. 
We have also proposed a feature fusion search strategy in the proposed searching scheme, which verified the importance of global/local fusion structure for the SR. Moreover, our complexity-based penalty to the reward could reduce the network complexity, which enabled lightweight network architecture. Experiments show that the resulted DeCoNASNet yields higher performance in terms of PSNR vs. complexity among the recent handcrafted lightweight SR networks and other NAS-based ones.

% conference papers do not normally have an appendix

% use section* for acknowledgment
\section*{Acknowledgment}

This research was supported by Samsung Electronics Co., Ltd.

%The authors would like to thank...

% trigger a \newpage just before the given reference
% number - used to balance the columns on the last page
% adjust value as needed - may need to be readjusted if
% the document is modified later
%\IEEEtriggeratref{8}
% The "triggered" command can be changed if desired:
%\IEEEtriggercmd{\enlargethispage{-5in}}

% references section

% Generated by IEEEtran.bst, version: 1.12 (2007/01/11)

%
% <OR> manually copy in the resultant .bbl file
% set second argument of \begin to the number of references
% (used to reserve space for the reference number labels box)

% that's all folks
\end{document}